\pdfoutput=1
\PassOptionsToPackage{table,dvipsnames}{xcolor}
\documentclass[11pt]{article}

\usepackage[]{acl}

\usepackage{times}
\usepackage{latexsym}

\usepackage[T1]{fontenc}

\usepackage[utf8]{inputenc}

\usepackage{microtype}

\usepackage{inconsolata}

\usepackage{graphicx}
\usepackage{microtype}
\usepackage{amsfonts}
\usepackage{mathrsfs}
\usepackage{multirow}
\usepackage{amssymb}
\usepackage{bbding}
\usepackage{subfigure} 
\usepackage{CJKutf8}
\usepackage{colortbl}
\usepackage{xcolor} 
\usepackage{booktabs}
\usepackage{makecell}
\usepackage{calligra}
\usepackage{algorithm}

\usepackage{algorithmicx}
\usepackage{algpseudocode} 
\usepackage{amsmath} 
\title{An Effective Incorporating Heterogeneous Knowledge Curriculum Learning for Sequence Labeling}

\author{
  Xuemei Tang\textsuperscript{1},\quad
  Jun Wang\textsuperscript{2},\quad
  Qi Su\textsuperscript{3}\thanks{Corresponding authors.},\quad
  Chu-Ren Huang\textsuperscript{1},\quad
  and Jinghang Gu\textsuperscript{1}\footnotemark[1]\\
  \textsuperscript{1}The Hong Kong Polytechnic University, Hong Kong, China \\
  \textsuperscript{2}Department of Information Management, Peking University, Beijing, China \\
  \textsuperscript{3}School of Foreign Languages, Peking University, Beijing, China \\
  \texttt{xuemeitang00@gmail.com}, \quad \texttt{\{sukia,junwang\}@pku.edu.cn} \\
  \texttt{\{churen.huang, jinghang.gu\}@polyu.edu.hk}
}

\begin{document}
\maketitle

\begin{abstract}
Sequence labeling models often benefit from incorporating external knowledge. However, this practice introduces data heterogeneity and complicates the model with additional modules, leading to increased expenses for training a high-performing model. To address this challenge, we propose a dual-stage curriculum learning (DCL) framework specifically designed for sequence labeling tasks. The DCL framework enhances training by gradually introducing data instances from easy to hard. 
Additionally, we introduce a dynamic metric for evaluating the difficulty levels of sequence labeling tasks. 
Experiments on several sequence labeling datasets show that our model enhances performance and accelerates training, mitigating the slow training issue of complex models~\footnote{\url{https://github.com/tangxuemei1995/DCL4SeqLabeling}}.
\end{abstract}

\section{Introduction and Related Work}

Sequence labeling is a core task in natural language processing (NLP) that involves assigning labels to individual elements in a sequence. Recent advancements in neural network methods have significantly improved performance in sequence labeling tasks ~\cite{em:44, Chen2017AFN, Zhang_Yu_Fu_2018, em:31, Nguyen_2021, Hou_Zhou2021, Liu_Fu_Zhang_Xiao_2021}. Some studies have explored integrating external knowledge, such as n-grams, lexicons, and syntax, to enhance these models. However, this integration adds heterogeneity and complexity to the input data. Additionally, incorporating such knowledge often necessitates extra encoding modules, like attention mechanisms ~\cite{Liu_Fu_Zhang_Xiao_2021, em:30} or graph neural networks (GNN) \cite{chen2017dagbasedlongshorttermmemory, Gui_Zou_Zhang_Peng_Fu_Wei_Huang, Nie_Zhang_Peng_Yang_2022}, which increase model parameters and make the system more computationally expensive to develop.

Curriculum Learning (CL)~\citep{Bengio} effectively addresses these challenges by simulating the human learning process, where training samples are introduced progressively from easy to hard. This approach facilitates efficient learning from heterogeneous data while enhancing both the speed and performance of the model~\cite{Bengio, Wang_Chen_Zhu_2021}. CL has shown success in a variety of NLP tasks, including machine translation~\cite{Wan_Yang_Wong_Zhou_Chao_Zhang_Chen_2020}, dialogue generation~\cite{Zhu_Chen_Wu_Liu_Zhao_2021}, and text classification \cite{Zhang_Wang_Cheng_Xiao_2022}.
Data-selection strategies are crucial in CL.
 However, these difficulty metrics primarily focus on the sentence level, such as~\citet{Mohiuddin_Koehn_Chaudhary_Cross_Bhosale_Joty_2022},  ~\citet{Yuan_Yang_Liang_Li_Liu_Huang_Xiao_2022} and ~\citet{NEURIPS2024_4ac4365b}'s works, and there is a lack of token-level and word-level metrics to measure the difficulty of sequence labeling tasks.

To address this gap, in this paper, we introduce a dual-stage curriculum learning (DCL) framework specifically designed for sequence labeling tasks. The first stage is data-level CL, where we train a basic teacher model on all available training data, aiming to alleviate the cold start problem of the student model. The second stage is model-level CL, where we start training the student model on a selected subset of the teacher model and gradually expand the training subset by considering the difficulty of the data and the state of the student model. Furthermore, we explore different difficulty metrics for sequence labeling tasks within the DCL framework. These metrics include a pre-defined metric, such as sentence length, and model-aware metrics, namely Top-N least confidence (TLC), Maximum normalized log-probability (MNLP), and Bayesian uncertainty (BU). Finally, we choose the classical sequence labeling tasks, Chinese word segmentation (CWS), part-of-speech (POS) tagging, and named entity recognition (NER), to validate our proposed approach.

\section{Method}

The framework proposed in this study consists of three main components: a teacher sequence labeling model, a student sequence labeling model, and a DCL training strategy. It is worth noting that the DCL is independent of the sequence labeling model. 

Following previous works \cite{Zhang_Yu_Fu_2018,em:41,em:39}, in sequence labeling tasks, we feed an input sentence $X =\{x_1,...x_{i}...,x_{M}\} $ into the encoder, and the decoder then outputs a label sequence $Y^* = \{y_1^*,...y_{i}^*...y_{M}^*\} $, where $y_i^*$ represents a label from a pre-defined label set $T$, and $M$ denotes the length of sentence.

\begin{figure*}[t]
    \centering
    \includegraphics[width=11.5cm,height=3.8cm]{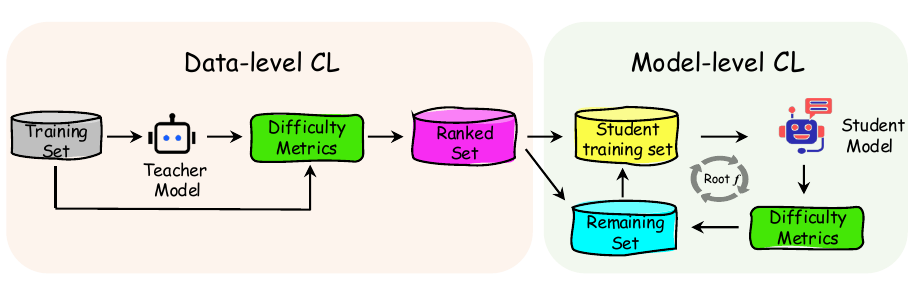}
    \caption{The framework of the proposed model consists of a teacher model, a student model, and a DCL strategy. Here, ``Root \textit{f}'' represents \textit{Root function}.}
    \label{f2}
    \vspace{-6mm}
\end{figure*}

\subsection{Dual-stage Curriculum Learning}

\begin{algorithm}[ht!]
\small
\renewcommand{\algorithmicrequire}{\textbf{Input:}}
\renewcommand{\algorithmicensure}{\textbf{Output:}}
\caption{Training Process with DCL}  
\label{a1}
\begin{algorithmic}[1]
    \Require Original corpus $\mathcal{D}$, difficulty metric $S(\cdot)$, teacher model epochs $E_0$, student model epochs $E_s$, scheduler $\lambda$, length function $|\cdot|$
    \Ensure Trained student model $\theta$
    
    \Statex \textit{// Data-level Curriculum Learning}
    \State Train teacher model $\theta_0$ on $\mathcal{D}$ for $E_0$ epochs
    \State Compute $S(\theta_0)$ for each sample in $\mathcal{D}$
    \State Sort $\mathcal{D}$ by $S(\theta_0)$ in ascending order to obtain ranked dataset $\mathcal{D}_r$
    
    \Statex \textit{// Model-level Curriculum Learning}
    \State Initialize $\lambda_0$ (starting curriculum ratio)
    \State $m \gets \lambda_0 \cdot |\mathcal{D}|$
    \State Student training set $\mathcal{D}_s \gets \mathcal{D}_r[0:m]$
    \State Remaining data $\mathcal{D}_o \gets \mathcal{D}_r[m:]$
    
    \For{$\textit{epoch} = 1$ to $E_s$}
        \If{$\lambda < 1$}
            \State \textit{a)} Train student model on $\mathcal{D}_s$ to obtain current $\theta_*$
            \State \textit{b)} Compute $S(\theta_*)$ for all samples in $\mathcal{D}_o$
            \State \textit{c)} Sort $\mathcal{D}_o$ by $S(\theta_*)$ in ascending order to get updated $\mathcal{D}_r$
            \State \textit{d)} Update $\lambda$ using Eq.~\ref{root}
            \State \textit{e)} Calculate new data size: $m \gets \lambda \cdot |\mathcal{D}| - |\mathcal{D}_s|$
            \State \textit{f)} Expand $\mathcal{D}_s$ with new samples: $\mathcal{D}_s \mathrel{+}= \mathcal{D}_r[0:m]$
            \State \textit{g)} Update remaining data: $\mathcal{D}_o \gets \mathcal{D}_r[m:]$
        \Else
            \State Train student model on $\mathcal{D}_s$
        \EndIf
    \EndFor
\end{algorithmic}
\end{algorithm}

We propose a novel dual-stage curriculum learning approach: \textit{data-level} CL and \textit{model-level} CL, as detailed in Algorithm~\ref{a1}.

At the data level, we first train a basic teacher model on the entire dataset $\mathcal{D}$ for $E_0$ epochs, where $E_0$ is smaller than the total epochs needed for convergence (Line 1). The teacher model $\theta_{0}$ is then used to calculate difficulty scores $S(\theta_{0})$ for each sample (Line 2), and the samples are sorted by difficulty to form a ranked dataset $\mathcal{D}_r$ (Line 3).

At the model level, we address the cold-start issue by initializing the student model training set $\mathcal{D}_s$ with a subset of $\mathcal{D}_r$ (Lines 4-6). The proportion of samples, controlled by the parameter $\lambda$, governs the curriculum learning process. The remaining data, $\mathcal{D}_o$, is incorporated into $\mathcal{D}_s$ gradually as $\lambda$ increases. The number of new samples to be added is denoted as $m$ (Lines 5, 14).

The student model is trained on $\mathcal{D}_s$ to update the model parameters $\theta_{*}$ (Line 10). Then, $\theta_{*}$ is used for the difficulty calculation of the samples in $\mathcal{D}_o$ (Line 11). Next, $\mathcal{D}_o$ is ranked by new difficulty scores, forming a new ranked dataset $\mathcal{D}_r$ (Line 12). 
The threshold $\lambda$ is updated (Line 13), and new samples are added to $\mathcal{D}_s$ based on $\lambda$ (Lines 14-15). As $\lambda$ approaches 1, all of $\mathcal{D}_o$ is added to $\mathcal{D}_s$. The complete dataset is then used to train the student model to convergence.

The key elements in Algorithm~\ref{a1} are the difficulty metric $S(\cdot)$ and threshold $\lambda$, which control the difficulty ranking of samples and the progression of training, respectively. The design of these components will be discussed in the following sections.

\subsection{Difficulty Metrics}

We now provide a detailed formulation for calculating the difficulty $S(\cdot)$ in Algorithm \ref{a1}.
In sequence labeling tasks, sample difficulty is tied to individual tokens, but assessing token-level difficulty is challenging. We use uncertainty from active learning to measure the model's confidence in labeling training samples.

\textbf{Bayesian Uncertainty (BU).} Following \citet{Buntine1991BayesianB}, model uncertainty can be assessed using Bayesian Neural Networks. As noted by \citet{Wang_Liu_Wang_Luan_Sun_2019}, higher predicted probability variance indicates greater uncertainty, suggesting that the model is less confident about the sample. In this work, we employ the widely-used Monte Carlo dropout \cite{Gal_Ghahramani_2016} to approximate Bayesian inference.

First, we apply Monte Carlo dropout \cite{Gal_Ghahramani_2016} to obtain each sample of token-level tagging probabilities. Specifically, for each token $x_i$, we perform $K$ stochastic forward passes through the model, resulting in $K$ predicted distributions ${P(y_i \mid x_i)}_{1}, \ldots, {P(y_i \mid x_i)}_{K}$. This provides $K$ predictions with associated probabilities for each token.
Then the expectation of token-level tagging probability can be approximated by
\begin{equation}
\resizebox{.70\hsize}{!}{$ \mathbb{E} [P(y_i|x_i)] \approx \frac{1}{K} \sum_{k=1}^{K} {P(y_i|x_i)}_{k} $}
 \vspace{-1mm}
\end{equation}
The variance of token-level tagging probability on the label set can be approximated by
\begin{equation}
\begin{aligned}
&var(x_i,\theta) \approx  \\
&\resizebox{.85\hsize}{!}{$\sum_{y_i\in T} (\frac{1}{K} \sum_{k=1}^{K} {P(y_i|x_i)}_{k}^2 - \mathbb{E} [P(y_i|x_i)]^2)$}
\end{aligned}
\end{equation}
Now, we obtain the variance of each token $var(x_i,\theta) $, then we use the average variance score of all tokens in the sequence as the sentence-level variance as follows.
\begin{equation}
var(\theta)_{aver.} = \frac{1}{M}\sum_{i=1}^{M} var(x_i,\theta) 
\end{equation}
The maximum variance score $var(\theta)_{max}$ also is valuable, which reflects the highest uncertainty in the sequence.
\begin{equation}
var(\theta)_{max}  = max_{i\in[1, M]} var(x_i,\theta)
\end{equation}

The final uncertainty score or difficulty score of each sequence is calculated as follows.
\begin{equation}
\label{e5}
S(\theta)^{BU} = var(\theta)_{max} + var(\theta)_{aver.}
\end{equation}

Both at the data level and model level, the difficulty of training samples is measured by the above various $S(\theta)$.

\subsection{Training Scheduler} 
\label{traing_s}
The training scheduler regulates the pace of CL. In our approach, we employ the \textit{Root function} as the control mechanism. This function ensures that the model receives sufficient time to learn newly introduced examples while gradually decreasing the number of newly added examples throughout the training process.
\begin{equation} 
\label{root} \lambda = \min\left(1, \sqrt{\frac{1 - \lambda_0^2}{E_{grow}} \cdot t + \lambda_0^2}\right)
\end{equation}
where $E_{grow}$ denotes the number of epochs required for $\lambda$ to reach 1, while $\lambda_0 > 0$ represents the initial proportion of the easiest training samples. $t$ indicates the $t_{th}$ training epochs. When $\lambda$ reaches 1, the model has access to the entire training dataset.

\begin{table*}[t]
\centering
\setlength{\tabcolsep}{1mm}
\begin{tabular}{c|c|cc|cc|cc}
\hline
\Xhline{1.2pt}
\multicolumn{1}{c|}{\multirow{2}{*}{\textbf{Model}}} &\multicolumn{1}{c|}{\multirow{2}{*}{\textbf{CL Setting}}}& \multicolumn{2}{c}{\textbf{CTB5}} & \multicolumn{2}{c}{\textbf{CTB6}}   &\multicolumn{2}{c}{\textbf{PKU}}  \\
\cline{3-8}
  &      & \multicolumn{1}{c}{CWS} & \multicolumn{1}{c}{POS}                & \multicolumn{1}{c}{CWS} & \multicolumn{1}{c}{POS}                                                       & \multicolumn{1}{c}{CWS} & \multicolumn{1}{c}{POS}               \\
\hline
\Xhline{1.2pt}

 \cline{2-8}

 \citet{em:31}  &-                  &    98.73 & 96.60         &   97.30 &         94.74                                             &   -  &    -                            \\

\citet{em:30} (McASP)&-     &  98.77         & 96.77    &  97.43    &   94.82     &  -               & -                  \\

\citet{Liu_Fu_Zhang_Xiao_2021}   &-             &  -        & 97.14      &   -  &         -                   &      -        &   -                           \\

\citet{Tang_Wang_Su_2024} (SynSemGCN)& -&  98.83  &  96.77             &   97.86  &    94.98     &  98.05    &   95.50       \\
\hline

 \multirow{5}{*}{McASP} &  Rand.  &   98.81  & 96.84& 97.37&94.90  &98.38&96.27\\ 

 &  Length & 98.83    & 96.85 &97.35 & 94.82  &98.40&  96.25         \\ 

 &  TLC  &  98.83   & 96.89& 97.37&94.83 &98.41& 96.30     \\ 
 
 &  MNLP   &   98.85  & 96.81& 97.41& 94.92&98.41& 96.30  \\ 

&  \cellcolor{gray!30} BU   &  \cellcolor{gray!30} \textbf{98.91}   &\cellcolor{gray!30} 96.87 & \cellcolor{gray!30}97.42&
\cellcolor{gray!30} 94.90 &\cellcolor{gray!30} 98.43&  \cellcolor{gray!30} 96.32                     \\ 
\hline

  \multirow{5}{*}{SynSemGCN}& Rand.   &  98.84 &  97.86 &       97.99        &95.05 & 98.48  &   96.40       \\

 & Length    & 98.80 &    96.84& 97.40         & 94.94 &            98.53         &  96.48  \\

 &TLC    &98.83  &  97.81  &        97.98      &  95.02             & \textbf{98.61} & \textbf{96.55}    \\            

 &MNLP    & 98.78 & 97.72  &  98.04&        95.13         & 98.56 &  96.48 \\             

 & \cellcolor{gray!30} BU  & \cellcolor{gray!30} 98.90 & \cellcolor{gray!30} \textbf{97.95} &  \cellcolor{gray!30} \textbf{98.05}   & \cellcolor{gray!30} \textbf{95.14} &   \cellcolor{gray!30}  98.59 & \cellcolor{gray!30} 96.54\\

\hline
\Xhline{1.2pt}
\end{tabular}
\caption{Experimental results of different models using different CL settings on test sets of three datasets.
Here, ``CWS'' represents the F1 value of CWS, and ``POS'' means the F1 value of the joint CWS and POS tagging. ``-'' means without the CL training strategy, and ``TLC'', ``MNLP'', and ``BU'' means using the DCL setting with different difficulty metrics. The maximum F1 scores for each dataset are highlighted.}
 \vspace{-3mm}
\label{t2}
\end{table*}

\section{Experiments}
\subsection{Dataset and Experimental Configurations}

\textbf{Dataset.} Chinese word segmentation (CWS) and part-of-speech (POS) tagging are representative sequence labeling tasks. So we evaluate our approach using three CWS and POS tagging datasets, including Chinese Penn Treebank version 5.0~\footnote{\url{https://catalog.ldc.upenn.edu/LDC2005T01}}, 6.0~\footnote{\url{https://catalog.ldc.upenn.edu/LDC2007T36}}, and PKU. More dataset details can be found in Appendix~\ref{dataset}.

\textbf{Teacher and student models.}
In this study, the basic transfer teacher framework is RoBERTa + Softmax. For the student model, we select two representative complex models introduced by \citet{em:30} and \citet{Tang_Wang_Su_2024}. In their work, \citet{em:30} employed an attention mechanism framework, McASP, to integrate lexicons and n-grams for the joint CWS and POS tagging task, using BERT as the encoder. Meanwhile, \citet{Tang_Wang_Su_2024} incorporated syntax and semantic knowledge into sequence labeling tasks through a GCN framework called SynSemGCN, with RoBERTa as the sequence encoder.

\textbf{Curriculum learning baselines}. We compare our difficulty metric with four baseline difficulty metrics for CL: \textit{\textbf{a. Random}}: Samples are assigned in random order; \textit{\textbf{b. Sentence Length (Length)}}: Samples are ranked from shortest to longest, based on the intuition that longer sequences are more challenging to encode; (Random and Length metics represent simple CL, namely without the teacher model). \textit{\textbf{c. Top-N Least Confidence (TLC)}}: The difficulty of a sequence is determined by using the $N$ tokens with the lowest confidence; \textit{\textbf{d. Maximum Normalized Log-Probability (MNLP)}}: The difficulty is assessed by calculating the product of the label probabilities for all tokens in the sequence.
The detailed computation processes for TLC and MNLP are provided in Appendix~\ref{Curriculum_Learning_Baselines}.

For further details on the important hyper-parameters of the model, please refer to Appendix~\ref{parameters_table}.
We discuss the selection process of these parameter values in detail in the Appendix~\ref{parameters}.

\subsection{Overall Experimental Results}
\label{s4.2}
Table~\ref{t2} presents the experimental results of baselines and two models with different CL settings. 
The experimental results reveal several noteworthy conclusions.

Firstly, the DCL methodology introduced in this paper is flexible and can be integrated with various complex models. As shown in Table~\ref{t2}, the difficulty metrics proposed here outperform the Random and Length metrics across most datasets. Specifically, the BU metric consistently delivers the best performance on the majority of datasets when applied to the SynSemGCN model, surpassing the TLC and MNLP metrics.

Additionally, we compare our approach with previous methods that incorporate external knowledge or resources into the encoder. The results reveal that models using CL exhibit significant performance improvements, surpassing the performance of earlier methods.

\begin{table}[H]
\centering

\begin{tabular}{lcc|c}
\hline
\Xhline{1.2pt}
 \multicolumn{1}{c}{\multirow{2}{*}{\textbf{Model}}} & \multicolumn{2}{c|}{\textbf{CTB5}}                                                &   \multirow{2}{*}{\textbf{Time}}                    \\ 
\cline{2-3}                                         
 & CWS &POS &  \\ 
\hline
Ours       &    \textbf{98.90}  & \textbf{97.95} &     287m         \\
\hline
w/o   data CL(BU)            &     \textbf{98.90}           &  97.88  &-                                   \\ 
\hline
w/o model CL(BU)  &    98.85        &97.51                     &   -                                                      \\ 
\hline
w/o  DCL     &   98.75               &   96.73                                 &  393m   \\ 

\hline
\Xhline{1.2pt}
\end{tabular}
\caption{Ablation experimental results of DCL. The baseline model ``w/o DCL'' denotes the model SynSemGCN; ``w/o model CL'' means the student model always uses the initial data order sorted by the transfer teacher model; ``w/o data CL'' indicates the initial training samples for the student model is drawn randomly from the training set; ``Ours'' indicates ``SynSemGCN+DCL(BU)''. Both teacher and student models with DCL in this table use BU as the difficulty metric. ``Time'' means the training time (in minutes).}
\label{ablation}
\end{table}

\begin{table}[H]
\centering
\begin{tabular}{ccc}
\hline
\Xhline{1.2pt}
\textbf{Model} & \multicolumn{2}{c}{\textbf{CTB5}} \\                  \cline{2-3}                         & CWS &POS  \\
\hline
McASP with BU  & \textbf{98.91} & \textbf{96.87} \\
\quad w/o $var(\theta)_{max}$ & 98.78 & 96.78 \\
\quad w/o $var(\theta)_{aver.}$ & 98.86 & 96.74 \\
McASP & 98.73 & 96.60 \\
\hline
\Xhline{1.2pt}
\end{tabular}
\caption{Ablation experimental results of two parts in BU metrics (Eq.~\ref{e5}).}
\label{tab:ctb5_results}
\end{table}

\begin{table*}[ht]
\centering
\small
\setlength{\tabcolsep}{0.8mm}
\begin{tabular}{lccc}
\toprule
\textbf{Models} & \textbf{Weibo (Chinese)} & \textbf{Note4 (Chinese)} & \textbf{CoNLL-2003 (English)} \\
\midrule
BERT & 66.22 & 79.15 & 90.94 \\
BERT + CL (Length) & 66.81 & 79.63 & 90.79 \\
BERT + DCL (TLC) & \textbf{67.52} & 79.53 & 91.30 \\
BERT + DCL (MNLP) & 65.73 & 79.95 & 91.15 \\
BERT + DCL (BU) & 66.74 & \textbf{80.02} & \textbf{91.77} \\
\hline
\Xhline{1.2pt}
\end{tabular}
\caption{Performance comparison of different difficulty metrics on three NER datasets.}
\vspace{-3mm}
\label{ner}
\end{table*}

\begin{figure}[b]
\includegraphics*[width=7cm,height=5cm]{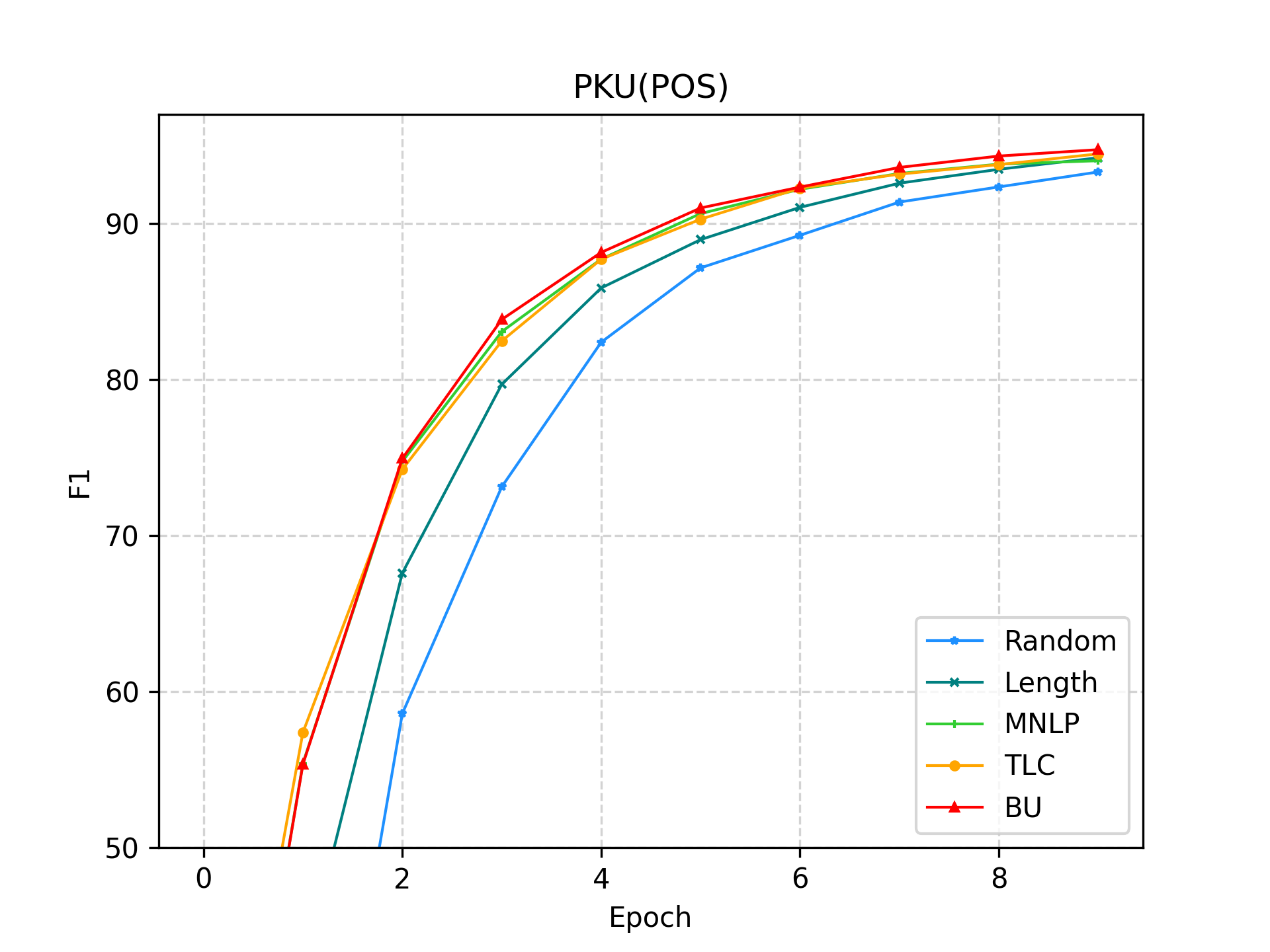}
\caption{The F1 scores on the dev set of PKU with different difficulty metrics in the model-level CL training process.  }
\label{f3}
\vspace{-8mm}
\end{figure}

\subsection{Effect of Dual-stage Curriculum Learning}
In this section, we discuss the impact of DCL. We perform ablation studies by removing either the data-level CL or the model-level CL. The results are summarized in Table~\ref{ablation}.
Model-level CL has a more significant impact than data-level CL. This is intuitive, as model-level CL influences the entire training process, while data-level CL primarily affects the early stages of student model training.

We also compare the training time of models with and without DCL. The experimental results in Table~\ref{ablation} show that all models were trained for 50 epochs. The training time for models using DCL includes the time spent on training the teacher model and calculating the difficulty values for the student model.
The results indicate that DCL improves model performance and reduces training time by over 25\%.

\subsection{Ablation Study on BU Difficulty Components}
We adopt McASP~\cite{em:30} as the backbone model, incorporating DCL as the training strategy and BU as the difficulty metric. To examine the contribution of each component in BU, we conduct ablation experiments on its two parts: $var(\theta)_{max}$ and $var(\theta)_{aver.}$, as shown in Table~\ref{tab:ctb5_results}. Removing either component results in performance degradation, indicating that both components are crucial. Moreover, the comparable drop in performance suggests that $var(\theta)_{max}$ and $var(\theta)_{aver.}$ contribute similarly to the effectiveness of DCL.

\subsection{Comparison of Difficulty Metrics}

In this section, we examine the impact of different difficulty metrics during the model-level CL training process for the SynSemGCN model. Figure~\ref{f3} shows the F1 score change on the PKU dataset development set over the first 10 epochs of model-level CL training. After 10 epochs, all training data are used, so the initial 10 epochs highlight the effect of different metrics. From the figure, we observe that BU, in particular, achieves the best performance, indicating that uncertainty-based metrics can select samples that better align with the model's learning trajectory, leading to faster learning.

\subsection{Generalization Capability}
We conduct additional experiments to demonstrate the applicability of our method to the NER task. We select two Chinese and one English NER datasets: Weibo~\footnote{\url{https://catalog.ldc.upenn.edu/LDC2013T19/}}, OntoNotes4~\footnote{\url{https://github.com/cchen-nlp/weiboNER}} and CoNLL-2003~\cite{tjong-kim-sang-de-meulder-2003-introduction}. 
The statistics of three datasets are shown in Table~\ref{ner_data}.
We compare the performance of models using DCL and CL (Length) with a model without CL on these datasets. As shown in Table~\ref{ner}, the results of the DCL method outperform those of BERT+CL (Length) and BERT (no CL), indicating the effectiveness of our method. This also suggests that our method can be applied to sequence labeling tasks beyond CWS and POS tagging.

\section{Conclusion}

This paper introduces a novel dual-stage curriculum learning framework aimed at enhancing performance and accelerating the training process for sequence labeling tasks. Focusing on the sequence labeling task of CWS, POS tagging, and NER, this framework demonstrates its effectiveness.

\section*{Limitations}
There are several limitations to our study. First, the design of our difficulty metrics involves the tuning of multiple hyperparameters, which may complicate optimization. Second, we did not explore a curriculum learning process that progresses from hard to easy examples. Third, we focused on a single variation of the $\lambda$ parameter to control CL and did not investigate alternative methods for adding training data.

\section*{Acknowledgments}
This research is supported by the NSFC project ``The Construction of the Knowledge Graph for the History of Chinese Confucianism'' (Grant No. 72010107003) and The Hong Kong Polytechnic University Project ``Evaluating the Syntax-Semantics Knowledge in Large Language Models'' (Grant No. P0055270).

\bibliography{custom}

\appendix

\section{Dataset}
\label{dataset}
The details of the three datasets are given in Table~\ref{t1}. Regarding the CTB datasets, we follow the same approach as previous works \cite{em:17,em:31} by splitting the data into train/dev/test sets. In the case of PKU, We randomly select 10\% of the training data to create the development set. 

\begin{table}[h]
\small
\centering
\setlength{\tabcolsep}{1mm}{\begin{tabular}{c|c|c|c|c}
\hline
\Xhline{1.2pt}
\multicolumn{2}{c|}{\textbf{Datasets}}  & \textbf{CTB5}  &\textbf{CTB6}  & \textbf{PKU}    \\
\hline
\multirow{2}{*}{Train} & Sent. & 18k  &23K   & 17k                       \\
\cline{2-5}
& Word & 494k & 99k & 482k  \\
\hline
\multirow{2}{*}{Dev}   & Sent. & 350 &2K  & 1.9k                      \\
\cline{2-5}
& Word & 7k & 60K & 53k    \\
\hline
\multirow{2}{*}{Test}  & Sent. & 348 &3K  & 3.6k                       \\
\cline{2-5}
& Word & 8k   & 12k & 97k  \\
\hline
\Xhline{1.2pt}
\end{tabular}}
\caption{Detail of the three datasets.}
\label{t1}
\vspace{-8mm}
\end{table}

\begin{table}[H]
\centering
\small
\setlength{\tabcolsep}{0.8mm}
\begin{tabular}{c|c|c|c|c} 
\Xhline{1.2pt}
\hline
\textbf{Datasets }                  & \textbf{Type}      & \textbf{Train} & \textbf{Dev}   & \textbf{Test} \\
\hline
\multirow{2}{*}{Weibo}     & Sentences & 1.35K & 0.27K & 0.27K \\
\cline{2-5}
 & Entities  & 1.89K & 0.39K & 0.42K                        \\ 
\hline
\multirow{2}{*}{OntoNotes} & Sentences & 15.7K & 4.3K  & 4.3K  \\
\cline{2-5}
& Entities  & 13.4K & 6.95K & 7.7K                     \\
\hline
\multirow{2}{*}{CoNLL2003} & Sentences & 15.0K & 3.5K  & 3.7K \\
\cline{2-5}
& Entities  &  23.5K& 5.9K & 5.7K                        \\
\hline
\Xhline{1.2pt}
\end{tabular}
\caption{Detail of the two NER datasets.}
\label{ner_data}
 \vspace{-4mm}
\end{table}

\section{Difficulty Metric Baselines}
\label{Curriculum_Learning_Baselines}

\textbf{Top-N least confidence (TLC).}
\citet{Culotta_McCallum_2005} proposed a confidence-based strategy for sequence models called least confidence (LC). This approach sorts the samples in ascending order based on the probability of the most possible label predicted by the model.

The least confidence of each token is calculated as follows.

\begin{equation}
\phi_{(x_i,\theta)}^{LC} = 1 - max_{y_{i}\in T} P(y_{i}|x_i)
\end{equation}
where $x_i$ is the $i_{th}$ token in the input sentence, $\theta$ denotes model parameters, $y_{i}$ is a pre-defined label, $T$ represents the pre-defined label set. $max_{y_{i}\in T} P(y_{i}|x_i)$ aims to find the probability of the most possible label predicted by the model. The smaller $\phi_{(x_i,\theta)}^{LC}$ reflects the more confident the model is in predicting the label of $x_i$.

According to~\citet{Agrawal_Tripathi_Vardhan_2021}, the confidence level of a sentence in a sequence labeling task is typically determined based on a set of representative tokens. Therefore, we select the top $N$ tokens with the highest least confidence in the sentence and then use their average value as the difficulty score of the sentence. Finally, the TLC difficulty metric is formulated as follows.
\begin{equation}
\resizebox{.60\hsize}{!}{$S(\theta)^{TLC} = \frac{1}{N} \sum_{n=1}^{N} \phi_{(x_n,\theta)}^{LC}$}
\end{equation}

\textbf{Maximum normalized log-probability (MNLP).} \citet{Shen_Yun_Lipton_Kronrod_Anandkumar_2018} used MNLP as a confidence strategy to find the product of the maximum probabilities of each token, which is equivalent to taking the logarithm of each probability and summing them. Finally, it is normalized to obtain the confidence score of the sentence as follows.
\begin{equation}
\begin{aligned}
&\prod_{i=1}^{M} max_{y_{i}\in T} P(y_{i}|x_i)  
\Longleftrightarrow \\
&\sum_{i=1}^{M} log \{max_{y_{i}\in T} P(y_{i}|x_i)\}
\end{aligned}
\end{equation}
where $M$ is the length of the sentence.
The difficulty of a sentence decreases as the confidence level increases. To account for this relationship, we introduce a negative sign. Additionally, in order to reduce the impact of sentence length, we apply a 
 normalization operation. Finally, MNLP is formulated as follows.
\begin{equation}
\resizebox{.85\hsize}{!}{$S(\theta)^{MNLP} = - \frac{1}{M} \sum_{i=1}^{M} log \{max_{y_{i}\in T} P(y_{i}|x_i)\}$}
\end{equation}

\section{Parameters Setting}
\label{parameters_table}

The key experimental parameter settings are shown in Table~\ref{t6}.

\begin{table}[h]
\centering
\small
\begin{tabular}{l|l}
\hline
\textbf{Hyper-parameters} & Value\\
\hline

$E_0$ & 5\\
\hline
$E_s$ & 50\\
\hline
$\lambda_{0}$ & 0.3 \\
\hline
$E_{grow}$ & 10 \\
\hline
$K$ & 3 \\
\hline 
$N$ & 5 \\
\hline
\end{tabular}
\caption{Experiment hyper-parameters setting.}
\label{t6}
\end{table}

\section{Effect of Hyper-parameters}
\label{parameters}

\begin{table}
    \centering
    \small
    \setlength{\tabcolsep}{3mm}
    \begin{tabular}{c|cccc}
    \Xhline{1.2pt}
     \hline
      \multirow{2}{*}{\textbf{$E_{grow}$}} & \multicolumn{2}{c}{\textbf{CTB5}} &\multicolumn{2}{|c}{\textbf{PKU}}\\
      \cline{2-5}
         & CWS&    POS&    CWS&   POS     \\

        \hline
        5    & 98.88        &97.89    & 98.65 & \textbf{97.01}\\
        10   &    \textbf{99.06}     &\textbf{98.96}  & \textbf{98.77} &96.97 \\
        15   &  98.84       &97.69   & 98.70 &96.90\\
        \hline
        \Xhline{1.2pt}
    \end{tabular}
    \caption{The effect of $E_{grow}$ in Eq.~\ref{root}.}
    \vspace{-4mm}
    \label{Egrow}
\end{table}

\begin{table}[]
\setlength{\tabcolsep}{3mm}
    \centering
    \small
    \begin{tabular}{c|cc|cc}
    \Xhline{1.2pt}
        \hline
        \multirow{2}{*}{\textbf{$E_0$}} & \multicolumn{2}{c}{\textbf{CTB5}}  &\multicolumn{2}{c}{\textbf{PKU}} \\
        \cline{2-5}
         & CWS& POS  & CWS& POS\\
        \hline
        5    & \textbf{99.06}     &\textbf{98.96}     & \textbf{98.77}    & \textbf{96.97}\\
        10   &    98.98     &97.90    & 98.54    & 96.69\\
        15   &  98.73       &96.87    & 98.53   & 96.54\\
        \hline
        \Xhline{1.2pt}
    \end{tabular}
    \caption{The impact of the number of epochs of teacher model, $E_0$. }
    \label{E_0_value}
\end{table}

\begin{table}[ht]
\setlength{\tabcolsep}{2mm}
    \centering
    \small
    \begin{tabular}{c|cc|cc}
    \Xhline{1.2pt}
        \hline
        \multirow{2}{*}{Para.} & \multicolumn{2}{c}{\textbf{CTB6}} &  \multicolumn{2}{c}{\textbf{PKU}} \\
        \cline{2-5}
        & CWS& POS  & CWS& POS \\
        \hline
        $K$=2  & \textbf{98.17}         & 95.43 &   98.73 & 96.58\\
        $K$=3  & 98.10  &\textbf{95.59} &\textbf{98.77} &\textbf{96.97}\\
        $K$=4  &  98.09          &95.56&    98.69    & 96.38\\
        \hline
        \Xhline{1.2pt}
    \end{tabular}
    \caption{The effect of of $K$ times dropout in BU difficulty metric.}
    \label{K_value}
     \vspace{-4mm}
\end{table}

In this section, we explore the impact of the hyperparameters on the performance of DCL. The adjustment of the parameters is based on the SynSemGCN+DCL(BU) model. 

\begin{figure}[ht!]
    \centering
    \includegraphics[width=7cm,height=5cm]{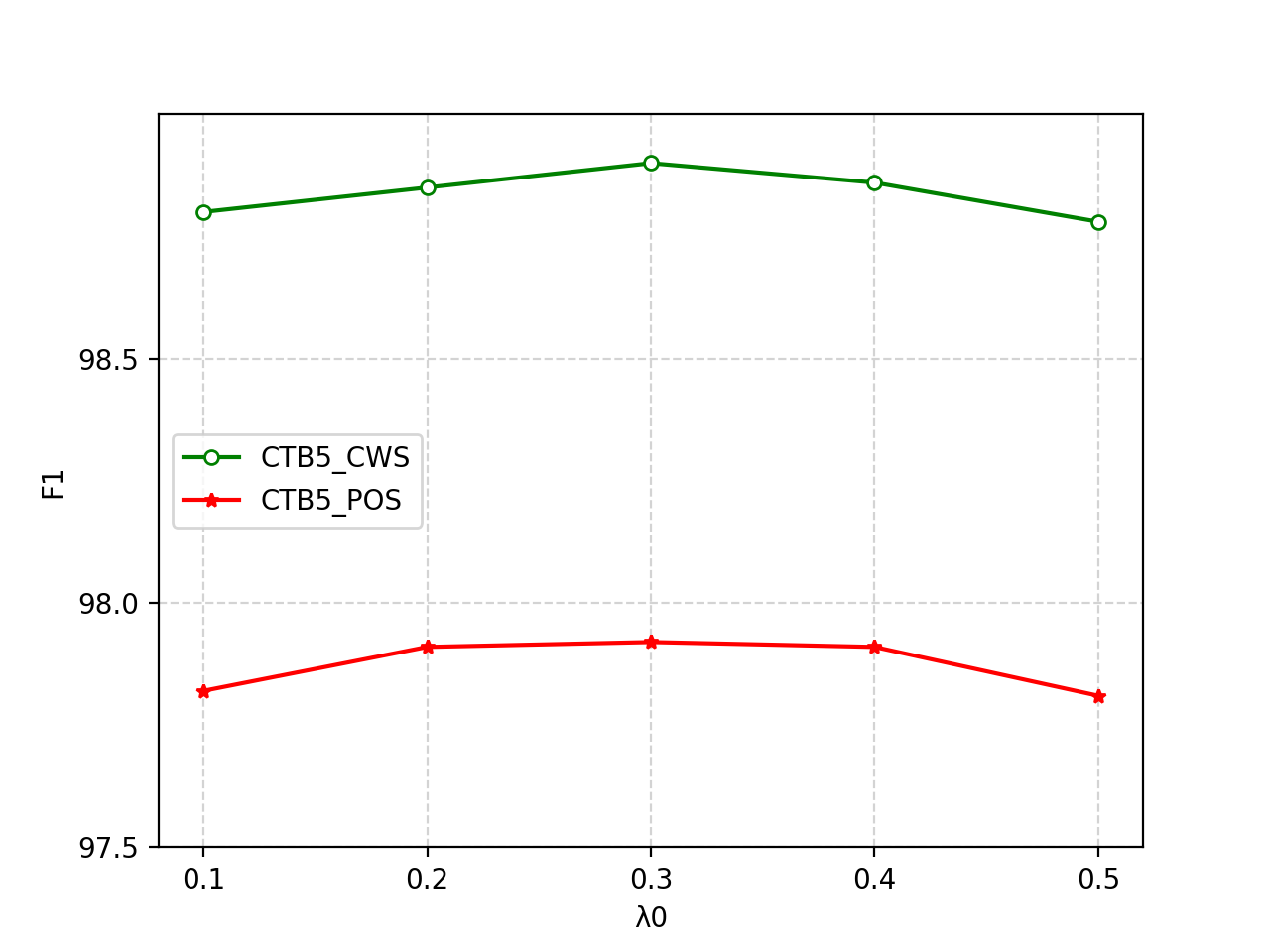}
    \caption{The impact of model-level curriculum learning hyper-parameters $\lambda_{0}$.}
    \label{lam}
 \vspace{-4mm}
\end{figure}

First, we investigate the impact of the hyper-parameter $\lambda_0$ on DCL performance. We conduct the experiments on the CTB5 dataset, tuning the value of $\lambda_0$ in the model-level pacing function Eq.~\ref{root}, and the experimental results are represented by a line graph as shown in Figure~\ref{lam}.
As observed, the model achieves optimal performance when $\lambda_0 = 0.3$. However, when the value exceeds 0.4, the model's performance gradually deteriorates.

Additionally, we examine the impact of $E_{grow}$ in Eq.~\ref{root}, which controls the number of epochs for $\lambda$ to reach 1. As shown in Table~\ref{Egrow}, when $E_{grow}$ is set to 10, the model exhibits superior performance on both the CTB5 and PKU datasets. Therefore, we adopt $E_{grow}$ as 10 epochs in our experiments.

Next, we assess the impact of the training epochs $E_0$ of the teacher model, which initializes the difficulty ranking of the training data for the student model. We aim to investigate whether a more mature teacher model contributes to improved performance. For this purpose, we conduct experiments on both the CTB5 and PKU datasets, utilizing teacher models trained for 5, 10, and 15 epochs to rank the initial training data for the student models. 

The experimental results, as shown in Table~\ref{E_0_value}, reveal that a more mature teacher model does not necessarily lead to better performance. Instead, the student model achieves optimal results when the teacher model is trained for 5 epochs. One possible explanation for this finding is that a teacher model with fewer training epochs aligns better with the initial state of the student model, allowing for a more suitable estimation of sample difficulty.

Then, we explore the impact of different $K$ values on the BU difficulty metric, which determines the number of dropout times. The experiments are conducted on the CTB6 dataset, and the results are summarized in Table~\ref{K_value}. Notably, the model achieves optimal performance when $K=3$. Therefore, we select $K=3$ for all the above experiments. 

Finally, we evaluate the effect of varying $N$ in the TLC metric. As shown in Table~\ref{N_value}, the best performance is achieved when $N=5$. 

\begin{table}[h]
    \centering
    \setlength{\tabcolsep}{2mm}
    \begin{tabular}{c|c|c}
    \hline
    \Xhline{1.2pt}
      \multirow{2}{*}{Para.} & \multicolumn{2}{c}{PKU}  \\
      \cline{2-3}
         & CWS &    POS     \\
        \hline
       $N$=1  & 98.55& 96.49\\
        \hline
        $N$=2  & 98.53 &96.49\\

        \hline
        $N$=3 & 98.56& 96.52\\

                \hline
       $N$=4  & 98.55& 96.51\\
        \hline
        $N$=5  & \textbf{98.71} &\textbf{96.64}\\

        \hline
        $N$=6 & 98.52  & 96.52 \\
                \hline
       $N$=7 & 98.56 & 96.51\\
        \hline
       $N$=8  & 98.55 & 96.48\\

        \hline
       $N$=9 & 98.53 & 96.49 \\
                \hline
      $N$=10 & 98.55 & 96.51\\
        \hline
       \Xhline{1.2pt}
    \end{tabular}
    \caption{The impact of $N$ in TLC difficulty metric. }
    \label{N_value}
    \vspace{-4mm}
\end{table}

\begin{table}[h]
\centering
\small
\setlength{\tabcolsep}{0.5mm}
\arrayrulecolor{black}
\begin{tabular}{c|c|c|c} 
\Xhline{1.2pt}
\hline
\multicolumn{2}{c|}{\textbf{Models}} & \multicolumn{2}{c}{\textbf{CTB5}}  \\ 
\hline
\textbf{A }& \textbf{B }                                                                 &\textbf{ CWS} & \textbf{POS} \\
\hline
BERT+DCL(BU)   & BERT                                         &  > &  >  \\ 
\hline
BERT+DCL(MNLP) & BERT                                        &  > &  >  \\ 
\hline
BERT+DCL(TLC)  & BERT                                            &  >&  >  \\ 
\hline
BERT+DCL(BU)   & BERT+CL(Length)                                      &  >&  >  \\ 
\hline
BERT+DCL(MNLP) & BERT+CL(Length)\textcolor[rgb]{0.753,0,0}{}               & > & > \\ 
\hline
BERT+DCL(TLC)  & BERT+CL(Length)                                        &  > &  >  \\
\Xhline{1.2pt}
\hline
\end{tabular}
\caption{Statistical significance test of F-score for our method and baselines on the CTB5 dataset.}
\label{ta1}
\vspace{-4mm}
\arrayrulecolor{black}
\end{table}

\section{Statistical Significance Test}
In this section, we conduct significance testing experiments. Following ~\citet{Wang_Zong_Su_2010}, we use the bootstrapping method proposed by ~\citet{Zhang_Vogel_Waibel_2004}, which is operated as follows.
In this process, starting with a test set $T_0$ comprising $N$ test examples, we repeatedly sample $N$ samples from $T_0$ to form $T_1$ and then repeat the process for $M$ times to form the test set collection, $\{T_1, T_2, ..., T_M\}$, where $M$ is set to 1000 in our testing procedure. Two systems denoted as $A$ and $B$, are assessed on the initial test set $T_0$, resulting in scores $a_0$ and $b_0$, respectively. The disparity between the two systems, labeled as $\delta_0$, is calculated as $\delta_0 = a_0 - b_0$. Repeating this process for each test set produces a set of $M$ discrepancy scores, denoted as $\{\delta_0, \delta_1, ..., \delta_M\}$. 

Following the methodology proposed by ~\citet{Zhang_Vogel_Waibel_2004}, we compute the 95\% confidence interval for the discrepancies (i.e., the 2.5th percentile and the 97.5th percentile) between the two models. If the confidence interval does not overlap with zero, it is affirmed that the differences between systems A and B are statistically significant ~\cite{Zhang_Vogel_Waibel_2004}.

Table~\ref{ta1} lists the significant differences between our system and the baseline system, where ``>'' indicates that the average value of $\delta$ exceeds zero, meaning that System A is better than System B; ``<'' indicates that the average value of $\delta$ does not exceed zero, meaning that System A is worse than System B; ``$\textasciitilde{}$'' indicates that there is no significant difference between the two systems. Finally, the comparison also indicates that our models are superior to the baseline.

\end{document}